\documentclass[10pt,twocolumn,letterpaper]{article}

\usepackage{iccv}
\usepackage{times,comment}
\usepackage{epsfig}
\usepackage{graphicx}
\usepackage{amsmath}
\usepackage{amssymb}
\usepackage{caption}
\usepackage{bm,paralist}
\usepackage[sort&compress,numbers]{natbib}
\usepackage{tabu,booktabs}
\usepackage{dsfont}

\def\ie{\emph{i.e}\onedot} 

\def\eg{\emph{e.g}\onedot} 

\def\etc{\emph{etc}\onedot}

\DeclareMathOperator*{\softmax}{Softmax}
\renewcommand{\vec}[1]{\mathbf{#1}}

\renewcommand{\AA}{\mathbf{A}}
\newcommand{\MM}{\mathbf{M}}
\newcommand{\ZZ}{\mathbf{Z}}
\newcommand{\UU}{\mathbf{U}}
\newcommand{\qq}{\mathbf{q}}
\newcommand{\vv}{\mathbf{v}}
\newcommand{\bsigma}{\boldsymbol{\sigma}}

\newcommand{\remove}[1]{}

\newcommand{\revise}[1]{{\color{black}{#1}}}

\usepackage[pagebackref=true,breaklinks=true,citecolor=blue,colorlinks,bookmarks=false]{hyperref}

\iccvfinalcopy 
\def\iccvPaperID{4320} %

\ificcvfinal\fi

\begin{document}

\title{Retrieve in Style: Unsupervised Facial Feature Transfer and Retrieval \vspace{-1.75ex}}

\author{Min Jin Chong$^{1}$\\
{\tt\small mchong6@illinois.edu}
\and
Wen-Sheng Chu$^{2}$\\
{\tt\small wschu@google.com}
\and
Abhishek Kumar$^{2}$\\
{\tt\small abhishk@google.com}
\and
David Forsyth$^{1}$\\ 
{\tt\small daf@illinois.edu}\vspace{-1.75ex}\\
\and
$^1${University of Illinois at Urbana-Champaign} \quad
$^2${Google Research} 
}

\twocolumn[{%
\renewcommand\twocolumn[1][]{#1}%
\maketitle
\vspace*{-6ex}
\begin{center}
    \centering
    \includegraphics[width=0.95\textwidth]{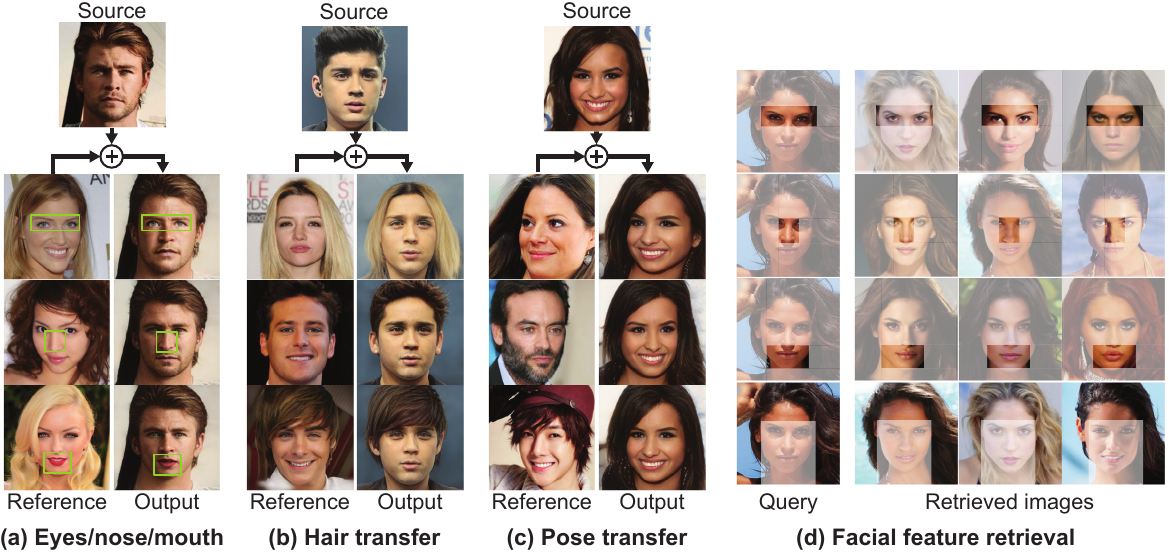}
    \vspace{-0pt}
    \captionof{figure}{
    We propose an {\em unsupervised} method to transfer local facial appearance from real reference images to a real source image, \eg, {\bf (a)} eyes, nose, and mouth. 
    Compared to the state-of-the-art \cite{collins2020editing}, our method enables photo-realistic transfers for {\bf (b)} hair and {\bf (c)} pose, and can be naturally extended for {\bf (d)} semantic retrieval according to different facial features. 
    }
    \label{fig:teaser}
    \vspace{0.2ex}
\end{center}%
}]

\maketitle

\begin{abstract}
\vspace{-8pt}
We present Retrieve in Style (RIS), an {\em unsupervised} framework for facial feature transfer and retrieval on real images. 
Recent work shows capabilities of transferring local facial features by capitalizing on the disentanglement property of the StyleGAN latent space.
RIS improves existing art on the following: 
1) Introducing more effective feature disentanglement to allow for challenging transfers (\ie, hair, pose) that were not shown possible in SoTA methods.
2) Eliminating the need for per-image hyperparameter tuning, and for computing a catalog over a large batch of images.
3) Enabling fine-grained face retrieval using disentangled facial features (\eg, eyes). To our best knowledge, this is the first work to retrieve face images at this fine level.
4) Demonstrating robust, natural editing on real images.
Our qualitative and quantitative analyses show RIS achieves both high-fidelity feature transfers and accurate fine-grained retrievals on real images. 
We also discuss the responsible applications of RIS. 
Our code is available at \url{https://github.com/mchong6/RetrieveInStyle}.
\vspace{-1ex}
\end{abstract}

\vspace{-2mm}
\section{Introduction}
\label{sec:intro}
\vspace{-1mm}

\remove{This paper shows how to disentangle the appearance of different facial features. Our method, Retrieve in Style (RIS) is able to control and edit individual facial features without any annotation of attributes. Detailed evaluation shows RIS ability to disentangle features leads to significant improvements in controlled generation. Furthermore, RIS leads to face image retrieval by feature -- one can search for all faces with the same nose, \etc. Evaluating this task then leads to a procedure evaluates how disentangled the representations are. }

\remove{Fine-grained face image editing and transfer using GANs remains an open-ended problem despite advancements in generating realistic high resolution images. %
Under unconditional settings, it is often hard to interpret or control the outputs of GANs. 
Conditional GANs are naturally more amenable thanks to extra annotations, such as manual labels \cite{bao2018towards,lample2017fader,xiao2018elegant,yin2017towards,choi2018stargan,zhang2018generative}, segmentation masks \cite{gu2019mask,lee2020maskgan}, attribute classifiers \cite{he2019attgan}, rendering models \cite{KowalskiECCV2020,usman2019puppetgan}, \etc. 
However, the degree of meaningful control over the generated images is largely dependent on how detailed the annotations are. 
This presents a challenge for fine-grained face editing, since detailed annotations often require additional computation and are difficult or impossible to acquire in the first place (\eg, the distinctive shape of eyes).}

Recent advancements in Generative Adversarial Networks (GANs) \cite{brock2018large,karras2019style,karras2020analyzing} have shown capabilities to generate realistic high resolution images, particularly for faces.
Under unconditional settings, it is often hard to interpret or control the outputs of GANs. 
Conditional GANs are more naturally amenable for semantic editing.
However, the degree of meaningful control over the output images is largely dependent on how detailed the annotations are. 
This presents a challenge for fine-grained face editing as it is often difficult or impossible to annotate datasets with the degree of detail needed for fine-grained editing.

Existing works on face editing typically leverage additional information to guide conditional generation, such as manual labels \cite{bao2018towards,lample2017fader,xiao2018elegant,yin2017towards,choi2018stargan,zhang2018generative}, segmentation masks \cite{gu2019mask,lee2020maskgan}, attribute classifiers \cite{he2019attgan}, rendering models \cite{KowalskiECCV2020,usman2019puppetgan}, \etc.
However, the additional information requires extra computation and is not always available in practice.
In addition, the fine-grained facial features (\eg, a distinctive shape of eyes) are difficult to describe as labels or features.
As an alternative, unsupervised discovery of latent directions in a pretrained GAN \cite{harkonen2020ganspace,voynov2020unsupervised,shen2020interpreting} allows for finding meaningful latent representations in a computationally efficient way. 
However, such approaches are less effective for fine-grained editing compared to supervised approaches. 

Recently, Editing in Style (EIS)~\cite{collins2020editing} proposed a mostly unsupervised method for facial feature transfer.  
While EIS allows semantic editing of spatially coherent facial features (\eg, eyes, nose and mouth), it requires computing a semantic catalog over the whole dataset and separate hyperparameter tuning for each image.
Such requirements make EIS non-scalable to large datasets as commonly encountered in retrieval domains.
In addition, it remains challenging for EIS to control facial features that are difficult to describe as a spatial map, such as hair and head pose.
More importantly, EIS works only on synthetic images and remains untested on how real images could be manipulated.

In this study, we propose Retrieve in Style (RIS), a simple and efficient unsupervised framework that tackles both fine-grained facial feature transfer and retrieval.
Fig.~\ref{fig:teaser} illustrates the capabilities offered by RIS.
RIS improves EIS in several aspects.
First, we discover the ``submembership'' property in the style space, showing that style channels corresponding to a particular feature (\eg, eyes) are different for every image and thus must be computed individually instead of over the entire dataset.
As the discovered channels are image-specific, RIS achieves more precise face editing for not only spatially coherent facial features (\eg, eyes, nose, mouth) but also challenging ones (\ie, hair, pose).
Second, with the discovered ``submembership'', we show it is possible to eliminate EIS's requirements on per-image semantic catalog and per-image hyperparameter tuning, and offer better scalability to larger problems.
Third, the image-specific representations naturally extend RIS for fine-grained facial feature retrieval that was not shown possible in EIS.
Lastly, we demonstrate that RIS offers editing and retrieval of {\em real images} when combined with GAN inversion methods, while EIS worked with synthetic images.
Although RIS is general and can be applied to a wide range of datasets, this study focuses on faces as there are established conventions on facial parts and its relevance in face retrieval applications (\eg, \cite{megreya2008matching,courtois1981target,bartlett1984typicality,lee2004suspect}).

\noindent
{\bf Our contributions are:}
\begin{compactenum}
    \item RIS improves over EIS based on our finding of ``sub-membership'', obtaining better controllability over facial features that are spatially coherent (eyes, nose, mouth) and incoherent (hair pose), while requiring no hyperparameter tuning.
    
    \item We obtain feature-specific representations (\eg, eyes, nose, mouth, hair), which enable face retrieval by fine-grained features that are difficult to describe or annotate even for humans. 
    To our best knowledge, this is the first work to address the fine-grained retrieval problem without supervision.
    
    \item  We show that RIS generalizes to GAN-inverted images, allowing transfer and retrieval on real images that was not shown possible in earlier studies. 
    Results on CelebA-HQ validates that RIS achieves high-quality retrieval on large, real-world datasets.
\end{compactenum}

\vspace{-2mm}
\section{Related Work}
\vspace{-1mm}

{\bf StyleGAN:}
StyleGAN1~\cite{karras2020analyzing} and StyleGAN2~\cite{karras2020analyzing} achieve state-of-the-art unconditional image generation. StyleGAN's unique architecture is inspired by style transfer work by Huang \etal~\cite{huang2017arbitrary}. Contrary to previous GAN architectures that map a random noise vector $\vec{z}$ to an image, StyleGAN maps $\vec{z}$ to $\vec{w} \in \mathcal{W}$ via a non-linear mapping network. Feature maps in the generator are then controlled by $\vec{w}$ in the AdaIN module~\cite{huang2017arbitrary}.

The $\mathcal{W}+$ latent space of StyleGAN has been shown to exhibit disentangled feature representations \cite{shen2020interpreting, karras2020analyzing,abdal2019image2stylegan,abdal2020image2stylegan++}. Xu \etal~\cite{xu2020generative} further showed that style coefficients $\bsigma$, where $\bsigma \!=\! FC(\vec{w})$ with 
$FC$ being an affine layer, demonstrate more disentangled visual features compared to $\vec{w}$. The style coefficients $\bsigma$ are directly used to scale the layer-wise activations in the generator.

{\bf Latent space image editing:}
Radford \etal~\cite{radford2015unsupervised} show that the latent space of GANs is semantically meaningful -- latent directions can be associated with semantics (\eg, pose, smile), with directions obtained by either supervised (\eg a pretrained attribute classifier, InterFaceGAN~\cite{shen2020interpreting}) or unsupervised means (\eg zooms and shifts,  Jahanian~\cite{jahanian2020on}).
Voynov~\cite{voynov2020unsupervised} finds directions corresponding to changes that can be observed by a classifer. GANSpace~\cite{harkonen2020ganspace} uses PCA to identify meaningful latent directions. Shen and Zhou \cite{shen2020closed} propose a closed-form factorization to obtain directions.

{\bf Feature activation image editing:} Local edits can follow from manipulating GAN feature activations. GAN Dissection~\cite{bau2018gan} uses a segmentation model to correspond internal GAN activations to semantic concepts, allowing them to add or remove objects. Feature Blending~\cite{suzuki2018spatially} recursively
blends feature activations between source and images to allow local semantics transfer. These methods require a pretrained segmentation model or user-provided masks.

One might obtain edits as image-to-image translations. AttGAN~\cite{he2019attgan} allows multi-attribute facial editing via a conditional GAN setup. StarGAN~\cite{choi2018stargan} proposes a single generator, multi-domain approach that uses conditional generation to achieve facial editing. GANimation~\cite{pumarola2018ganimation} conditions the generator with Action Units annotations to allow smooth facial expression editing. MaskGAN~\cite{lee2020maskgan} uses segmentation masks to enable interactive spatial image editing.

{\bf Face retrieval:}
Current facial retrieval systems generally match faces based on identities and lack the granularity to match on a facial feature level. Non deep-learning based retrieval systems such as Photobook~\cite{pentland1996photobook} and CAFIIRIS~\cite{wu1993facial} use features such as Eigenfaces~\cite{turk1991eigenfaces}, textual descriptions, and/or facial landmarks; but we expect learned features to have advantages. FaceNet~\cite{schroff2015facenet} learns embeddings via a triplet loss where the Euclidean distances between embeddings correspond to facial similarity by training with identities. Other works \cite{sun2015deeply, taigman2014deepface} formulate the problem as a classification task between identities. But these methods perform retrieval at the level of identity and by design, are invariant to details such as expressions and hairstyles. In contrast, RIS aims to improve the granularity of face retrieval. Instead of asking to ``retrieve faces with similar features'' we are asking to ''retrieve faces with similar eyes, nose, mouth, \etc``.

{\bf GAN Inversion:}
GAN inversion encodes a real image to the latent space of a GAN. It is commonly done via gradient descent in the latent space \cite{karras2020analyzing, abdal2020image2stylegan++, wulff2020improving} which leads to accurate reconstruction at the expense of scalability. An encoder-based approach~\cite{zhu2020domain, richardson2020encoding, xu2020generative} instead allows scalable GAN inversion.

\vspace{-2mm}
\section{Retrieve in Style}
\vspace{-2mm}
In this section, we describe the proposed Retrieve in Style (RIS) for both facial feature transfer and retrieval.
We first review Editing in Style (EIS) \cite{collins2020editing} that our method is built upon.
Then, we propose improvements to EIS for a more controllable and intuitive transfer, and show that our method can be naturally extended for fine-grained face retrieval, which was not possible in EIS.

\subsection{Editing in Style}

Unlike methods that manipulate the latent space via vector arithmetic \cite{jahanian2020on,shen2020closed,shen2020interpreting,voynov2020unsupervised,harkonen2020ganspace}, EIS formulates the semantic editing problem as copying style coefficients $\bsigma$ of StyleGAN \cite{karras2019style} from a reference image to a source image, \ie, the output image carries facial features from the reference images while preserving the remaining features from the source image. 
The authors show that semantic local transfer is possible on images generated by a pretrained StyleGAN with minimal supervision.

One key insight of EIS is that spatial feature activations of a StyleGAN generator can be grouped into clusters that correspond to semantically meaningful concepts such as eyes, nose, mouth, \etc. 
Specifically, let $\AA \in \mathbb{R}^{N \times C \times H \times W}$ be the activation tensor at a particular layer of StyleGAN, where $N$ is the number of images, $C$ the number of channels, $H$ the height and $W$ the width. 
Spherical $K$-way k-means~\cite{buchta2012spherical} is applied spatially over $\AA$, \ie, clustering over $N \times H \times W$ vectors of size $C$.
Each spatial location of $\AA$ is associated with cluster memberships $\UU \in \{0,1\}^{N \times K \times H\times W}$, and 
then used to compute a contribution score $\MM_{k,c}\in[0,1]^{K \times C}$:
\begin{equation}
\vspace{-2mm}
    \MM_{k,c} = \dfrac{1}{NHW} \sum_{n,h,w} \AA^2_{n,c,h,w} \odot \UU_{n,k,h,w}.
    \label{eq:1}
\end{equation}
Intuitively, $\MM_{k,c}$ tells how much the $c$-th channel of style coefficients $\bsigma\in\mathbb{R}^C$ contributes to the generation for facial feature $k$. Note that $\bsigma$ directly scales the activations $\AA$ in the modulation module --- the larger the activations, the more $k$ is affected by the channel $c$.

Transferring a facial feature $k$ across two images is then performed via interpolation between style coefficients $\bsigma^S,\bsigma^R$ of the source and the reference images. 
The style coefficient of the edited image $\bsigma^G_k$ can be obtained by rewriting the style interpolation in Eq.~(3) of \cite{collins2020editing}:
\begin{equation}
    \bsigma^G_k = (1-\qq_k) \odot \bsigma^S + \qq_k \odot \bsigma^R, \label{eq:interpolation}
\end{equation}
where $\qq_k\in[0,1]^C$ is the interpolation vector for a given facial feature $k$.
EIS finds $\qq_k$ using a greedy optimization derived from $\MM_{k,c}$ and manual hyperparameter tuning to determine which channels to ignore.
Such hyperparameters can be sensitive to different reference images and lead to suboptimal transfers, as shown in Sec.~\ref{sec:experiments}. 
In addition, $\MM_{k,c}$ is computed over $N$ images and is fixed for all feature transfers. 
We argue in Sec.~\ref{sec:method:transfer} that having a fixed $\MM_{k,c}$ may not be ideal for transfer, as not all images share the same channels to describe the same facial feature.

\begin{figure}[t]
    \centering
    \includegraphics[width=1\linewidth]{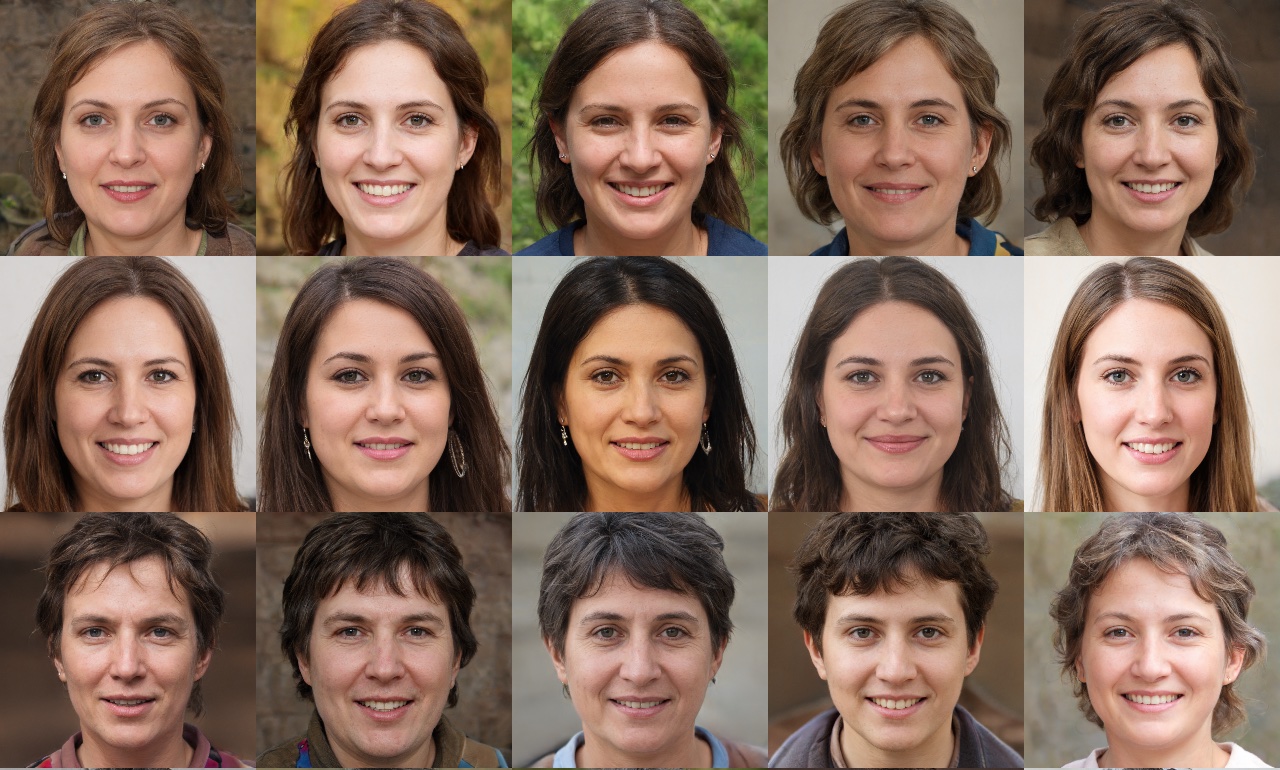}
    \vspace{-16pt}
    \caption{
    {\bf Submembership:}
    Contribution scores $\MM_{k}$ from our method allow meaningful clustering. In this figure, each row is a cluster for $k\!=\!\text{hair}$; images within a row are similar, showing that clustering is effective. Across rows, the images differ, showing that there is real variation in the hair.
    }
    \label{fig:hair_cluster}
\end{figure}

\subsection{Improving EIS for Facial Feature Transfer}
\label{sec:method:transfer}

{\bf Submemberships:}
EIS assumes that the channels that make a high contribution for a particular feature (say, eyes) are the same for each image.
So to compute $\MM_k$ in Eq.~\eqref{eq:1}, EIS averages the scores over a large collection of images of size $N$. 
We hypothesize the high-contribution channels may vary from image to image. 
This means averaging over $N$ images can lose details specific to the source or reference. 

We visualize the presence of this effect in Fig.~\ref{fig:hair_cluster}. %
Performing Spherical k-means clustering over per image $\MM_{\text{hair}}$ $(N=1)$ of images 
in a dataset yields semantically meaningful clusters. 
Images in each row belong to the same cluster. 
The hairstyles within the same row are similar, while hairstyles across rows are distinctively different. 
We further analyze the top active channels (each channel corresponds to a dimension of $\MM_{k}$) for each cluster, and observe that each cluster has its own set of top active channels that are unique to it. Please refer to supplementary materials for more detailed analyses.
This validates our hypothesis that high-contribution channels for a semantic feature are not the same across images. 
That is, the same feature $k$ of different images are controlled by different groups of channels.
We term these groups as ``submembership'', which is a crucial motivation for this work.

With ``submembership'' in mind, instead of computing $\MM_{k,c}$ over a large batch of $N$ images, we show that the responsible channels are more accurately computed over only the source and reference images, \ie, $N=2$. Specifically,
\begin{align}
\vspace{-3mm}
    \MM_{k,c} = \max \bigg(&\sum_{h,w} \AA[s]^2_{c,h,w} \odot \UU[s]_{k,h,w}, \nonumber\\ 
                           &\sum_{h,w} \AA[r]^2_{c,h,w} \odot \UU[r]_{k,h,w}\bigg ),
    \label{eq:2}
\vspace{-2mm}
\end{align}
where $s$ and $r$ indicate the particular source and reference images of interest, respectively. 
Intuitively, to transfer from a reference to a source image, we are interested in channels that are important to source, reference, or both.

\begin{figure}[t]
    \centering
    \includegraphics[width=1\linewidth]{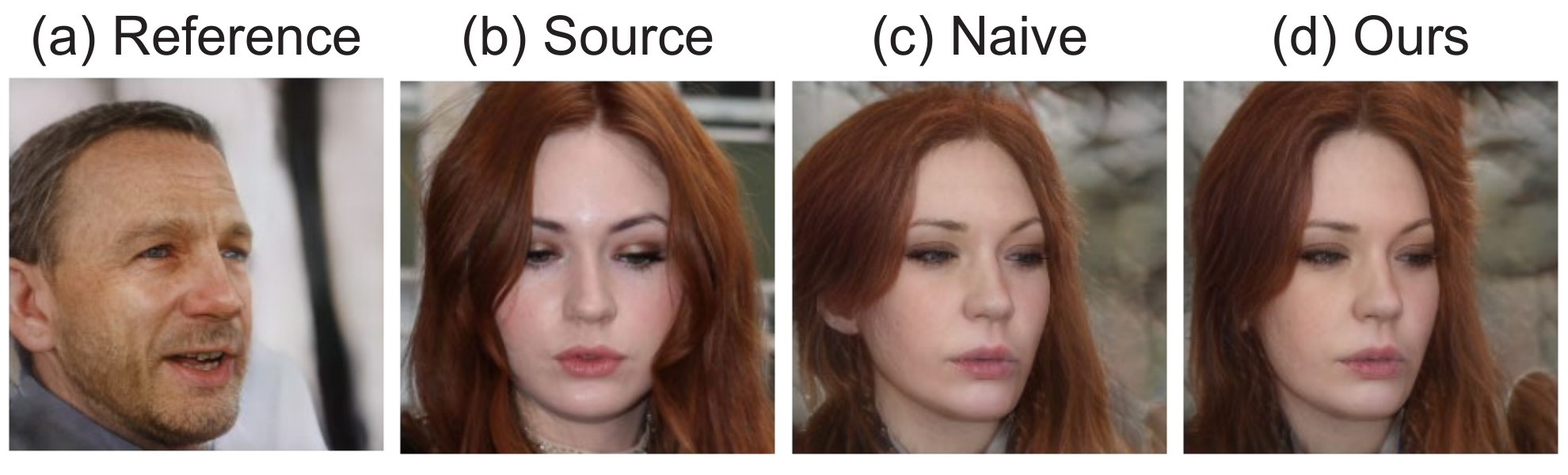}
    \vspace{-19pt}
    \caption{
    {\bf Pose transfer} from {\bf (a)} reference to {\bf (b)} source.
    {\bf (c)} Naively copying style coefficients from the first 4 layers of StyleGAN2~\cite{karras2020analyzing} transfers primarily pose and partially hair (shorter hair on left, flatter hair top), showing their style coefficients are entangled in the early layers. 
    {\bf (d)} Our method matches the pose of the reference image and preserves the hair faithfully from the source.
    }
    \label{fig:pose}
    \vspace{-1.5ex}
\end{figure}
\begin{figure*}[thb]
    \centering
    \includegraphics[width=1\linewidth]{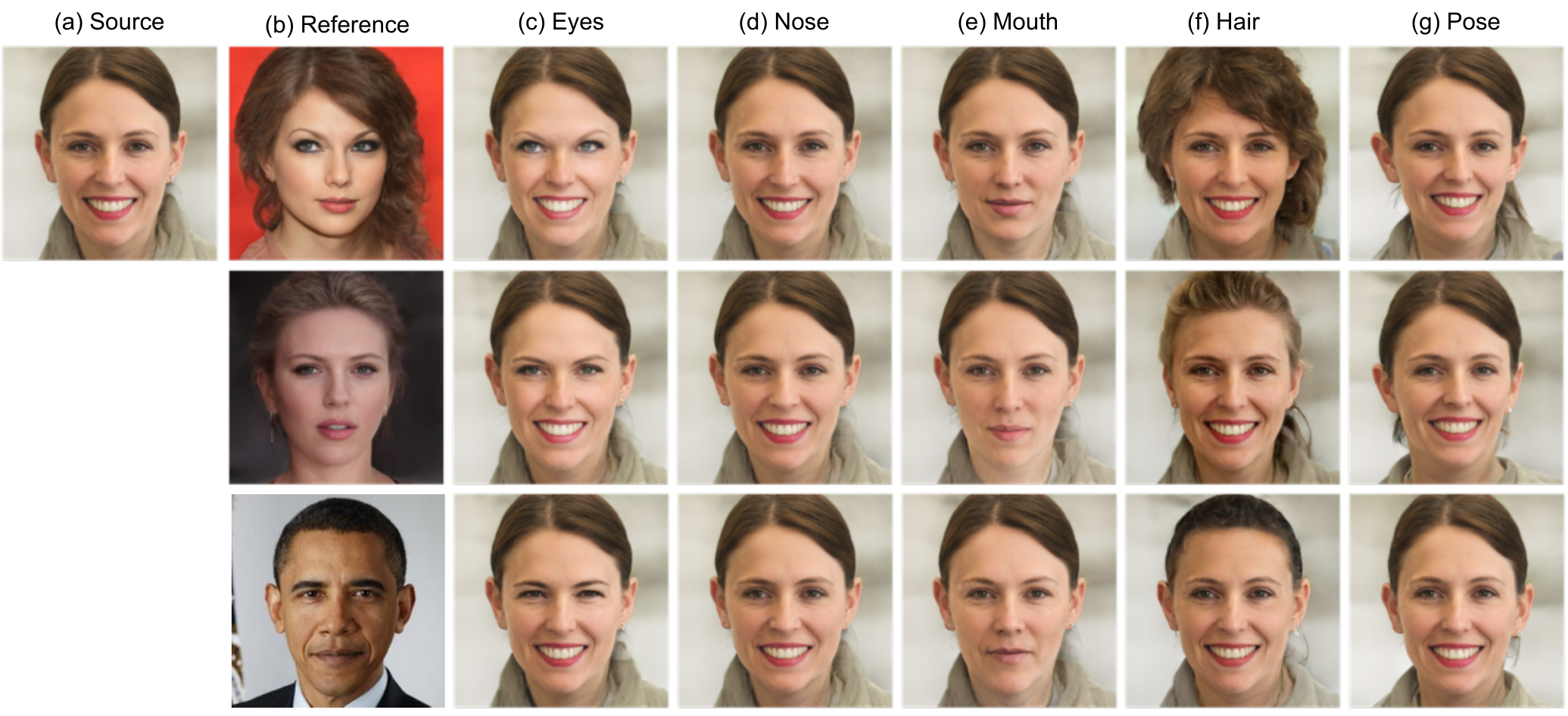}
    \vspace{-3ex}
    \caption{
    {\bf Facial feature transfer:}
    Our method performs effective semantic editing on real images by transferring facial features from {\bf (b)} a reference image to {\bf (a)} a source image. 
    Our method transfers spatially coherent features (\ie, eyes, nose, mouth) as well as challenging features hair and pose.
    Note that real image editing is not possible with SoTA EIS~\cite{collins2020editing}.
    }
    \label{fig:my_transfer}
    \vspace{-1.5ex}
\end{figure*}

\begin{figure}[t]
    \centering
    \includegraphics[width=1\linewidth]{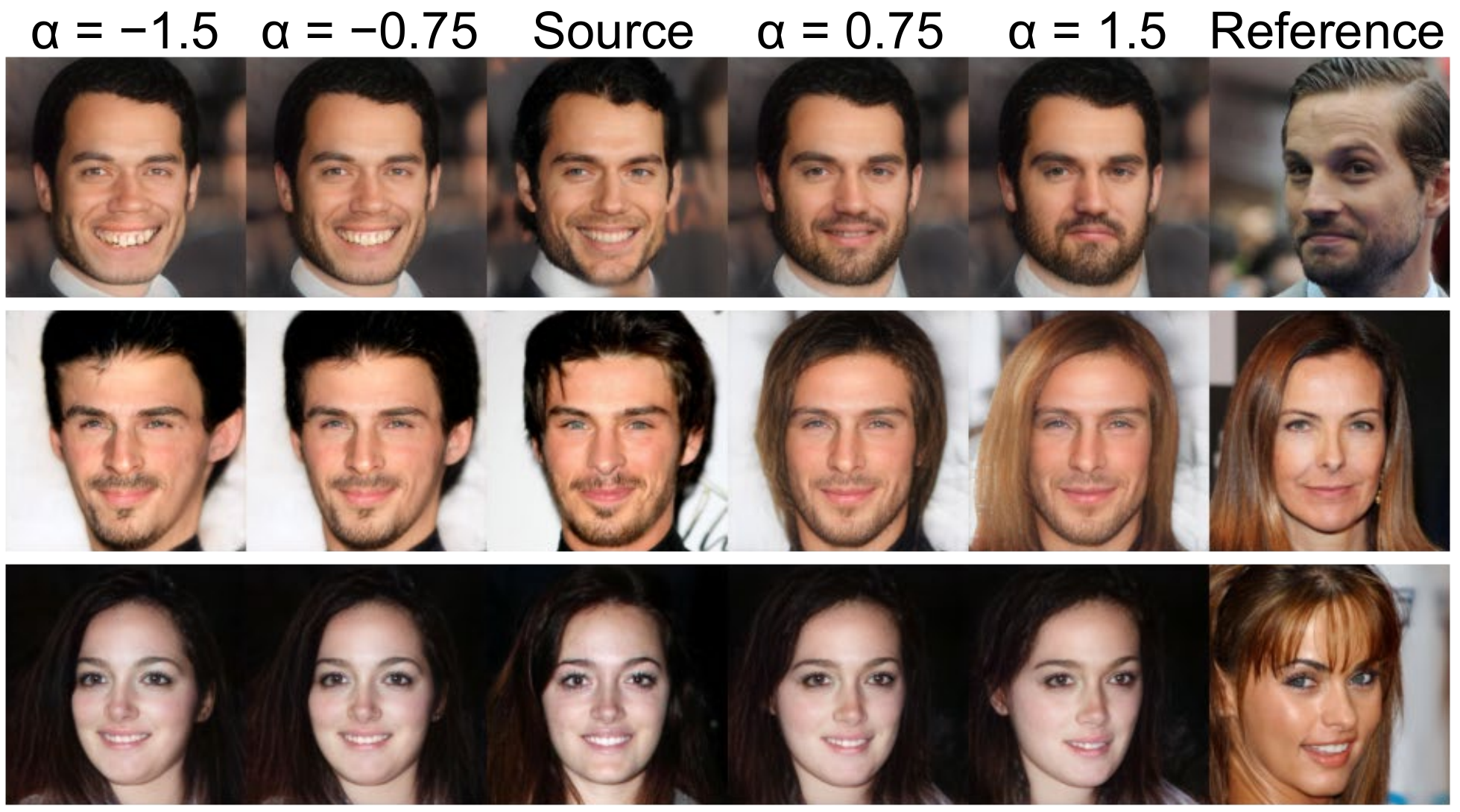}
    \vspace{-12pt}
    \caption{
    {\bf Latent direction:}
    The $\alpha$ variable in RIS controls interpolation between the source and the reference images, showing a smooth transition of mouth (top row), hair (middle row) and pose (bottom row).
    }
    \label{fig:alpha}
    \vspace{-2ex}
\end{figure}

{\bf Obtaining interpolation vector:}
Instead of getting the interpolation vector $\qq_k$ from the greedy optimization process (like in EIS) which is dependent on per-image hyperparameters $\rho$ and $\epsilon$, %
we assume each channel of the style coefficient $\bsigma$ corresponds to one facial feature. This follows from the disentangled style space of StyleGAN and in practice, works well. Under this assumption, we obtain a soft class assignment for each style coefficient channel with a softmax of all classes (rows of $\MM$), obtaining:
\begin{equation}
    \qq = \softmax_k \bigg(\frac{\MM}{\tau} \bigg),
\label{eq:q_softmax}
\end{equation}
where $\vec{M} \in [0,1]^{K \times C}$ is the stacked contribution score of all facial features, $\tau$ is the temperature, $\qq \in [0,1]^{K \times C}$ is the interpolation vector. The interpolation vector for a particular feature $k$, $\qq_k$ can be indexed from the row of $\qq$. $\qq_k$ can be thought of the mask for $k$ that allows interpolation between $\bsigma^S$ and $\bsigma^R$.

{\bf Pose transfer:}
Karras \etal~\cite{karras2020analyzing} have shown that the first few layers of StyleGAN2 capture high level features such as pose. In Fig.~\ref{fig:pose}, we show that copying the style coefficients of the first 4 layers of StyleGAN2 (which corresponds to the first 2048 style coefficient channels), transfers mostly pose and hair information from reference to source image, leaving other features like eyes and mouth untouched. By assuming that the first $4$ layers \emph{only} contain pose and hair information, we simply derive:
\begin{equation}
    \qq_{\text{pose}} = \vec{1} - \qq_{\text{hair}},
\end{equation}
for only the first $4$ layers with the rest zeroed out. Similarly, for all facial features other than hair, the first $4$ layers are zeroed out to prevent pose changes. 
As shown in Fig.~\ref{fig:pose}, $\qq_{\text{pose}}$ captures pose information without affecting hair.

One significant advantage of our pose transfer is that it requires no labels or manual tuning. 
For example, GANSpace~\cite{harkonen2020ganspace} requires manually choosing layer subsets; AttGAN~\cite{he2019attgan} and InterFaceGAN~\cite{shen2020interpreting} requires attribute labels, StyleRig~\cite{tewari2020stylerig} requires a 3D face model.
\revise{Fig.~\ref{fig:my_transfer} illustrates our full capability of facial feature transfer.}

{\bf Latent Direction:}
Unlike EIS that limits facial feature transfer to style interpolation as in Eq.~\eqref{eq:2}, we formulate the problem as traversing along the latent direction, based on work showing StyleGAN's latent space vector arithmetic property~\cite{radford2015unsupervised}. 
Then, we revise Eq.~\eqref{eq:2} to:
\begin{equation}
    \bsigma^G_k = \bsigma^S + \alpha\qq_k \odot (\bsigma^R-\bsigma^S), \label{eq:latent_dir} 
\end{equation}
where the latent direction is $\vec{n} = \qq_k \odot(\bsigma^R-\bsigma^S)$ and the scalar step size is $\alpha$. If we restrict $\alpha \in [0,1]$, we will be performing a style interpolation. Under the property of vector arithmetic, we can instead use $\alpha \in \mathcal{R}$ which allows style extrapolation. We show in Fig.~\ref{fig:alpha} that scaling $\alpha$ allows an increase or decrease in the particular facial property. For example, we are able to do smooth pose interpolation.

\begin{figure*}[thb]
    \begin{minipage}{0.31\textwidth}
        \begin{center}
            \includegraphics[width=3.2\textwidth]{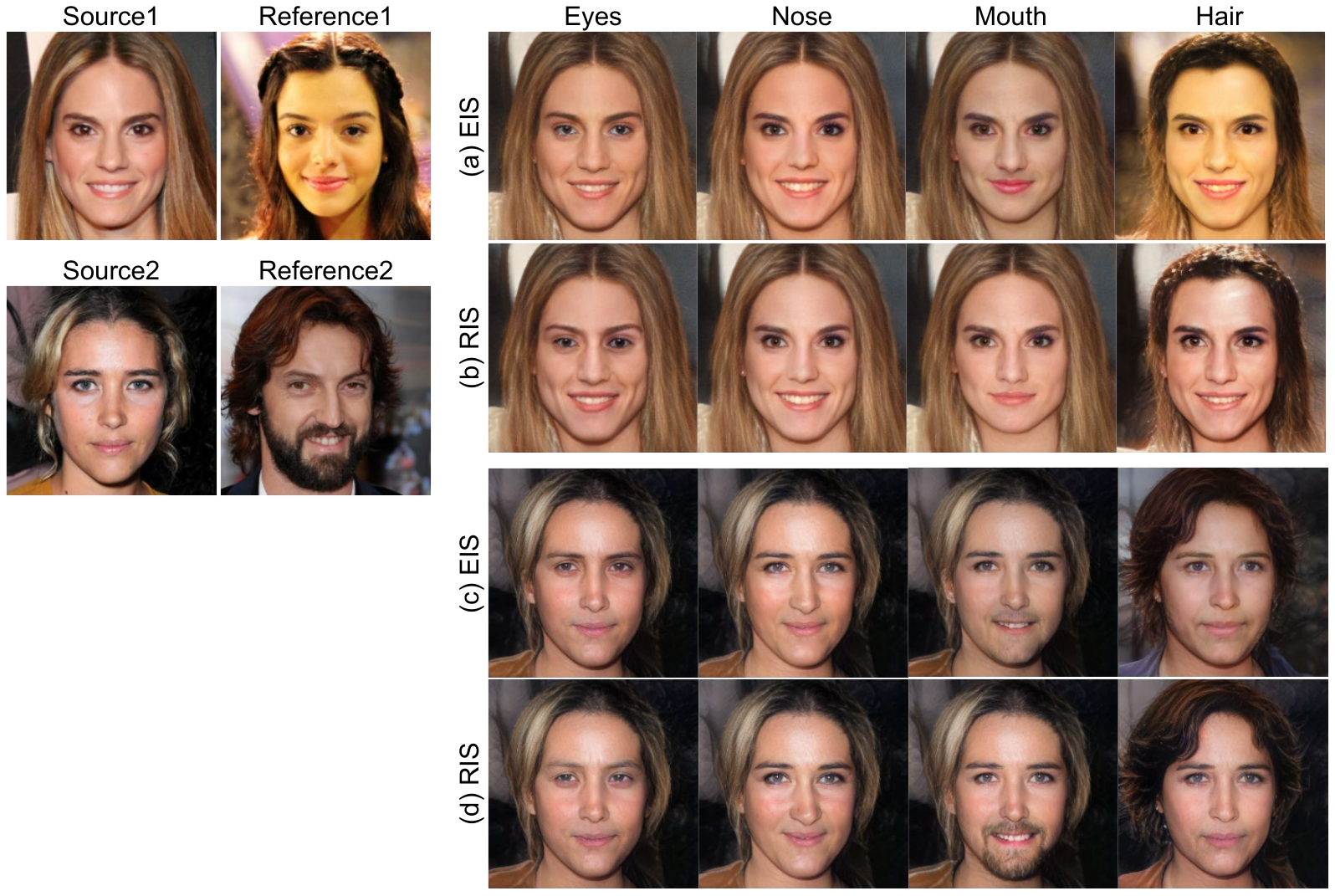}            
        \end{center}
        \vspace{-34ex}
        \caption{
        {\bf Comparison with EIS \cite{collins2020editing}:}
        {\bf (a, b)} show transfers from Reference 1 to the source image; {\bf (c)(d)} from Reference 2. 
        Our method (RIS) generates visually more accurate and natural results.
        {\em E.g.,} EIS changed the skin tone in (a) and shirt color in (c), while RIS does not. 
        RIS also achieves beard transfer around mouth in (d), even though beard on female faces is rare or absent in training data.
        }
        \label{fig:compare}
    \end{minipage}
\end{figure*}

\subsection{Facial Feature Retrieval}
\label{sec:method:retrieval}

This section shows the style representation in Eq.~\eqref{eq:latent_dir} can be adapted to fine-grained facial feature retrieval, which is defined as follows.
Given a query image $I_Q$ and a retrieval dataset $\mathcal{X}$, we aim to retrieve the top-K closest images $\mathcal{T}_k \subset \mathcal{X}$ with respect to a facial feature (\eg, eyes). 
As described in the previous section, RIS identifies the style channels that mediate the appearance of facial features for particular images. 
This suggests the style channels can be used to retrieve faces with appearance similar to the facial features in a query face. 
Face retrieval is usually done by matching on an identity embedding~\cite{schroff2015facenet,sun2015deeply,taigman2014deepface}. 
However, fine-grained facial feature retrieval is relatively unexplored as it is difficult to collect and annotate training data with fine granularity (\eg, shape of the eyes or nose).

For each facial feature $k$, we have $\qq_k \in [0,1]^{1 \times C}$ to encode, for a particular image, how much that channel contributes to that feature. Since $\qq_k$ can be considered as a mask, we construct a feature-specific representation:
\begin{equation}
    \vv^Q_k = \qq^Q_k \odot \bsigma^Q. 
    \label{eq:retrieval}
\end{equation}
Feature retrieval can be then performed by matching $\vv_k$, as two images with similar $\vv_k$ suggest a lookalike feature $k$. 

\begin{figure*}
    \includegraphics[width=1\linewidth]{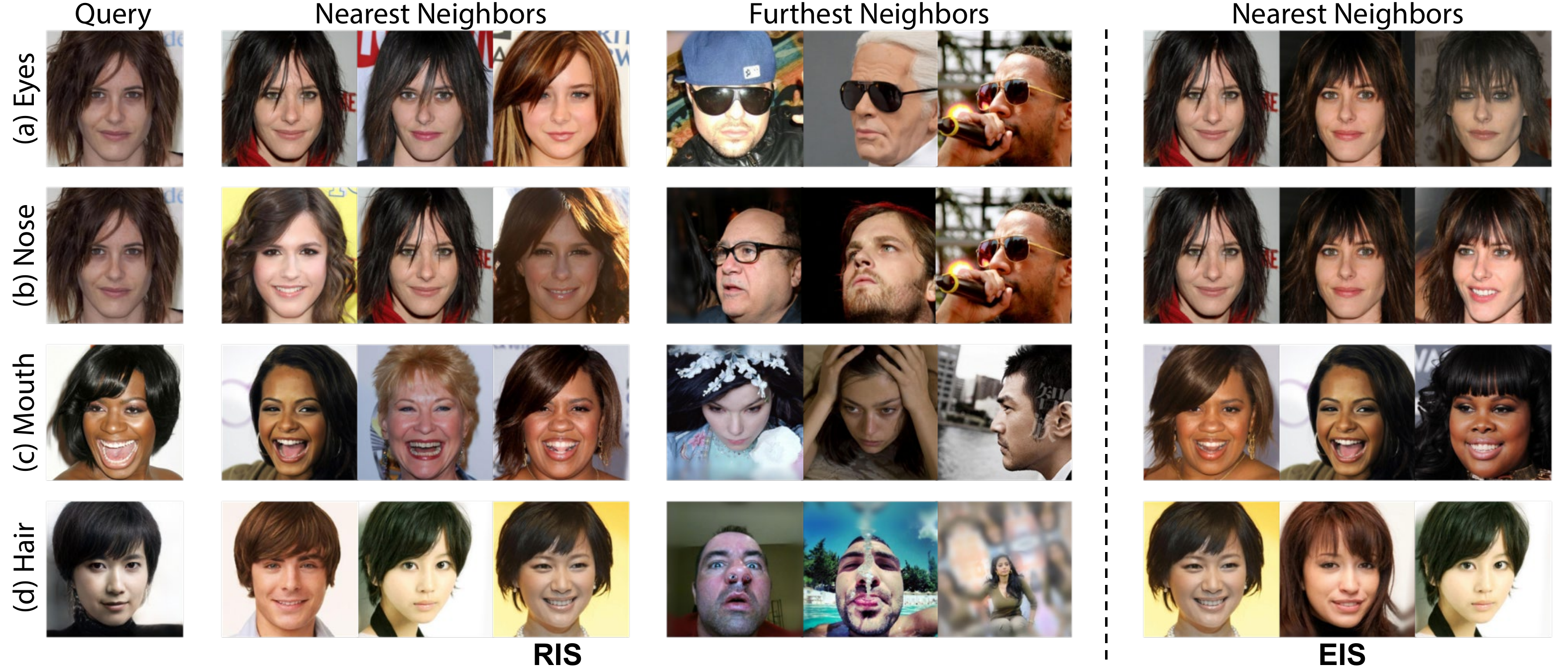}
    \vspace{-4ex}
    \caption{
    {\bf Facial feature retrieval:}
    We compare fine-grained retrieval between our method RIS (submembership $\MM_k$) and EIS \cite{collins2020editing} (universal $\MM_k$) on real faces. We show 3 faces each from nearest and furthest retrieval (NR and FR). RIS retrieves semantically similar NRs on all facial features while showing variance on non-matching features. Note EIS retrieves very similar NR on eyes and nose with same query image indicating a lack of feature localization.
    \vspace{-1ex}
    }
    \label{fig:nn_retrieve}
\end{figure*}

We compute the representations $\vv^R_k = \vec{q}^R_k \odot \bsigma^R$ where $\bsigma^R \in \boldsymbol{\Sigma}$ and $\boldsymbol{\Sigma}$ are the style coefficients for the images in $\mathcal{X}$. 
We then define the distance between the facial features of two style coefficients/face images as 
\begin{equation}
    \text{Distance}_k(I^Q, I^R) = d(\vv^Q_k, \vv^R_k),
    \label{eq:distance}
\end{equation}
where $d$ is a distance metric (cosine distance in this study). 
We then rank the distances for nearest neighbor search for facial feature $k$. Intuitively, if there is a $\MM_k$ and consequently, a $\qq_k$ mismatch between two images, their distance will be large. Since Fig.~\ref{fig:hair_cluster} shows that similar features have similar $\MM_k$, vice versa, it follows that smaller distances will reflect more similar features. We show this is true empirically and RIS works as in expectation from Fig.~\ref{fig:nn_retrieve}. Additionally, we observe better results if we normalize $\bsigma^Q$ and $\bsigma^R$ using layer-wise mean and standard deviation from $\boldsymbol{\Sigma}$.

{\bf Comparison between SoTA EIS \cite{collins2020editing} and RIS (ours):}
Both EIS and RIS share a unique way to perform unsupervised local face editing by attributing transfers to reference images.
They differ in how they accomplish it.
(1) EIS computes the contribution score $\MM$ by averaging over a batch of $N$ images.
Based on the findings of $\MM$'s submembership, RIS uses $N=2$, which avoids manual per-image hyperparameter tuning and thus allows a more scalable and intuitive transfer.
As a result, RIS yields more precise transfer of eyes, nose, and mouth, and enables transferring novel features such as hair and pose that were not shown possible in EIS.
(2) RIS redefines $\MM$ as an image-specific representation, which allows for unsupervised fine-grained face feature retrieval.
EIS assumes an averaged representation of $\MM$, which will be shown in experiments to be less effective for feature retrieval.

\section{Experiments}
\label{sec:experiments}
\vspace{-2pt}

While other work based on StyleGAN, including EIS~\cite{collins2020editing, harkonen2020ganspace}, focus on manipulating generated images, we focus on the more relevant problem of manipulating \emph{real images}. 
This is a more difficult problem as there are no guarantees that GANs performing well on generated images are stable enough to generalize to real images.

To show that RIS generalizes to real datasets, we use CelebA-HQ~\cite{karras2017progressive} with 30k images for all our experiments.
Since feature-based retrieval requires the inversion of the entire dataset, we opt to use pSp~\cite{richardson2020encoding}, a SoTA encoder-based GAN inversion method, for all our experiments. 

\subsection{Facial Feature Transfer}
\vspace{-2pt}

In this section, we provide qualitative and quantitative analyses for facial feature transfer on real images. We fixed $\tau \!=\! 0.1$ and $\alpha=1.3$ for all experiments, as we observed the temperature $\tau$ in Eq.~\eqref{eq:q_softmax} is insensitive to different source and reference images. We used $N\!=\!200$ for EIS \cite{collins2020editing} following the authors' implementation. %

{\bf Qualitative analysis:}
Fig.~\ref{fig:compare} shows a qualitative comparison between RIS (our method) and EIS on real images. 
It can be observed that RIS offers better localization ability. 
EIS (Fig. \ref{fig:compare}(a)) affects skin tone heavily across all transfers, notably changing lighting heavily for hair transfer. 
In contrast, RIS maintains relatively similar skin tones while transferring the targeted features. 
EIS also changes the eyes and nose of the source image while transferring mouth (Fig. \ref{fig:compare}(a)), indicating entanglement in their representations. 
While transferring mouth (which includes the chin region), EIS fails to reproduce the beard in the image Reference2 (Fig. \ref{fig:compare}(c)).
On the other hand, RIS faithfully reproduces the beard (Fig. \ref{fig:compare}(d)). 
It is noteworthy that RIS is able to generate a female face with beard, representing an out-of-distribution generation that is absent in the training set. 
Please refer to supplementary materials for more comparisons.

\begin{table}[t]
    \centering
    \begin{minipage}{0.22\textwidth}
        \begin{tabular}{lc}
            \toprule
            {\bf Method} & {\bf FID$_\infty$}  \\ 
            \midrule
            StyleGAN2 \cite{karras2020analyzing} & 2.44 \\
            EIS \cite{collins2020editing} & 3.47 \\
            RIS (ours) & 3.73 \\
            \bottomrule
        \end{tabular}
    \end{minipage}
    \hfill
    \begin{minipage}{0.23\textwidth}
        \vspace{-6pt}
        \includegraphics[width=\textwidth]{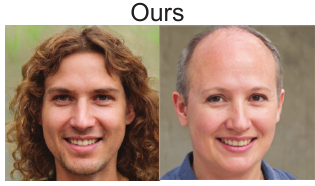}
    \end{minipage}
    \vspace{-4pt}
    \caption{
        {\bf Image fidelity comparison:} 
        RIS achieves a comparable FID$_\infty$ compared to EIS and is only slightly worse compared to the base StyleGAN2. The larger FID$_\infty$ can be attributed to our capability of OOD generation, \eg, long-hair males or bald females as in the right image.
    }
    \label{table:fid}
    \vspace{-2ex}
\end{table}

{\bf Quantitative analysis:}
To quantitatively validate our transfer results, we computed FID$_\infty$ \cite{chong2020effectively}, an unbiased estimate of FID, for baseline StyleGAN2 \cite{karras2020analyzing}, EIS \cite{collins2020editing} and RIS. Details on the setup are provided in the supplementary.

Table \ref{table:fid} shows the FID$_\infty$ comparison.
Both EIS and RIS achieved small FID$_\infty$ differences compared to the base StyleGAN2. However, RIS yielded a slightly larger FID$_\infty$, which can be explained by the ability of our method to even generate out-of-distribution samples, if needed for transferring features. Such samples are uncommon in the FFHQ dataset that trains the base StyleGAN2, and thus contribute to a larger FID$_\infty$, \eg, our method is capable of transferring long hair to a bearded male, or bald hair to a female, as shown in the right of Table \ref{table:fid}.

\vspace{-1mm}
\subsection{Facial Feature Retrieval}
\vspace{-1mm}
\label{sec:retrieval}

We evaluate our retrieval performance qualitatively and quantitatively. We use GAN inverted CelebA-HQ images as the retrieval dataset, and cosine distance as the metric.

{\bf Qualitative analysis:}
As fine-grained facial retrieval is relatively unexplored, to the best of our knowledge, there are no proper metrics to evaluate this task. 
Instead, we repurposed the averaged $\MM_k$ in EIS for retrieval and use it as a baseline.  
Specifically, when computing retrieval representations in Eq.~\eqref{eq:retrieval}, we replaced the individual $\qq_k$ with the $\qq_k$ derived from EIS's averaged $\MM_k$. 
Since large-scale hyperparameter tuning for every reference image is infeasible for EIS, we obtained $\qq_k$ with a fixed hyperparameter choice that may not generalize to all images. 

Fig.~\ref{fig:nn_retrieve} shows qualitative comparisons between RIS and EIS. 
RIS has observably more disentangled representations. 
Specifically, for eyes retrieval, although the query has distinct eyes, RIS retrieves images with the same eyes but different identities, while EIS only retrieves the same identity.
This suggests EIS representations are entangled between eyes and identity features.
In addition, EIS retrieves almost the same images for different features (\ie, eyes and nose), suggesting entanglement. 
For mouth retrieval, RIS recognizes the wide open mouth of the query, retrieving semantically similar (w.r.t.~the mouth feature) yet diverse images. 
EIS, on the other hand, retrieves images with the same skin tone, suggesting a lack of feature localization. 
Lastly, for hair retrieval, RIS retrieves images with similar hair but with different genders, while EIS only retrieves only female images. 
Finally, furthest neighbors for RIS differ semantically from the query image.

Overall, RIS nearest neighbors exhibit significant variance on non-matching features while EIS nearest neighbors do not. 
Along with our superior results in Fig.~\ref{fig:compare}, this further reinforces that our individual $\MM_k$ yields better disentanglement and feature focus compared to the averaged $\MM_k$ in EIS. 
This also validates our hypothesis of submemberships.

\begin{table}[]
\centering
\begin{tabular}{lcr}
\toprule
& \multicolumn{2}{c}{\bf Attribute Matching Score (\%)}          \\
                            \multicolumn{1}{l}{{\bf Class}}  & \multicolumn{1}{c}{\bf{Ours}}  & \multicolumn{1}{r}{{\bf EIS}}\\

\midrule
Eyes  & 96.3  & 95.4 \\
Nose  & 100.0 & 100.0 \\
Mouth & 81.1  & 75.8 \\
Hair  & 97.5  & 97.1 \\
\bottomrule
\end{tabular}
\caption{We compare AMS between RIS and EIS to measure retrieval accuracy {\em w.r.t.} a given facial feature using a pretrained attribute classifier. 
RIS outperforms EIS in all classes, with \emph{mouth} retrieval being noticeably better.}
\label{table:AMS_tbl}
\vspace{-1ex}
\end{table}

\begin{figure}[t]
    \centering
    \includegraphics[width=.7\linewidth]{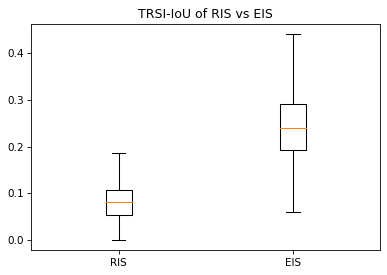}
    \caption{
    TRSI-IoU measures the extent of overlapping identities between two different feature queries on the same face. 
    Methods that disentangle facial features better are expected with smaller TRSI-IoU (see text). We compare a boxplot of TRSI-IoU for RIS and EIS. 
    RIS shows noticeable improvement in the median (red line) with much smaller interquartile range (boxes).
    This suggests our method better disentangles facial features.
    }
    \label{fig:iou}
\end{figure}

\textbf{TRSI-IoU.~}
We use retrieval to evaluate how well RIS disentangles facial features. 
We focus on two retrieved set identity IoU (TRSI-IoU): retrieve two sets of images using two facial feature queries on the same face;
TRSI-IoU is computed as intersection-over-union of the identities between these two sets.
A full face retrieval method should have a TRSI-IoU close to 1 if the two queries are the same person, and 0 otherwise. 
Assume a method does not disentangle features, it is possible to approximately predict (say) mouth from eyes. 
In turn, retrieving using eyes (resp.~mouth) will implicitly constrain mouth (resp. eyes), so the two retrieved sets will have many individuals in common;
hence TRSI-IoU becomes relatively large. 
On the other hand, if a method properly disentangled (say) eyes and mouth, its identities should not overlap much; thus TRSI-IoU becomes relatively small.
The minimum obtainable value of TRSI-IoU is difficult to know, but lower TRSI-IoU is good evidence a method  disentangles better. Fig~\ref{fig:iou} shows boxplots of TRSI-IoU for RIS and EIS, evaluated for 100 queries and all pairs of facial features (chosen from eyes, nose, mouth, hair). 
RIS shows significantly lower TRSI-IoU, and the difference is statistically significant.

\textbf{Attribute Matching Score.~}
We used attribute classifiers pretrained on CelebA attributes \cite{liu2015faceattributes} to further evaluate the quality of our retrieval. 
Note that these attributes are binary and not sufficiently detailed for fine-grained purposes.
There is also a distinct lack of diversity in CelebA and its attributes, \eg, lack of head coverings, curly hairs, \etc, which makes evaluation of RIS on generating faces of diverse and inclusive people not possible.

The intuition of our procedure is as follows: for retrieval of the $k$-th feature {\em hair}, hair-related attributes $\mathcal{A}_k$ (\eg, ``black\textunderscore hair'', ``wavy\textunderscore hair'', \etc) should remain similar between query and retrieved images. 
Please see supplementary for the full list of attributes associated with $k$. 

We retrieve top-5 images $\mathcal{T}_5^{(i)}$ according to a query image $I_Q^{(i)}$ for a feature $k$. 
We took an attribute classifier $\mathcal{F}$, and got its prediction for the $a$-th attribute as $\widehat{\mathcal{F}}_a(\cdot) = \left[\mathcal{F}_a(\cdot) >T \right]$, \ie, $\widehat{\mathcal{F}}_a(\cdot)\!=\!1$ if the prediction is larger than threshold $T=0.5$ and 0 otherwise. 
Then, Attribute Matching Score (AMS) is defined for the $k$-th facial feature:
\begin{align}
   \text{AMS}_k \!=\! \dfrac{ \sum_{I_Q^{(i)}\in\mathcal{X}} \sum_{t^{(i)}\in\mathcal{T}_5^{(i)}} \sum_{a\in\mathcal{A}_k} \left[\widehat{\mathcal{F}}_a(I_Q^{(i)}) \!=\! \widehat{\mathcal{F}}_a(t^{(i)}) \right]}{|\mathcal{X}| \cdot |\mathcal{T}_5^{(i)}| \cdot |\mathcal{A}_k|}.
   \nonumber
   \label{eq:AMS}
\end{align}
Table \ref{table:AMS_tbl}(b) compares AMS scores between EIS and RIS. As the classifier is trained on predefined attributes that do not contain fine granularity, it could be less descriptive to our particular task of fine-grained retrieval. Still, RIS outperforms EIS in all classes under this less-granular setting, with \emph{mouth} retrieval being noticeably better.

\vspace{-1mm}
\section{Conclusion}
\vspace{-1mm}
\label{sec:conclusion}
\vspace{-0ex}

We presented Retrieve in Style (RIS), a simple and efficient unsupervised method of facial feature transfer that works across both short-scale features (eyes, nose, mouth) and long-scale features (hair, pose) on real images without any hyperparameter tuning. 
RIS produces realistic, accurate feature transfers without modifying the rest of the image, and naturally extends to the fine-grained facial feature retrieval. 
Note that techniques for photorealistically manipulating images could be misused to produce fake or misleading information, and researchers should be aware of these risks.
To the best of our knowledge, this is the first work that enables unsupervised, fine-grained facial retrieval, especially so on real images. 
Our qualitative and quantitative analyses verify the effectiveness of RIS.

\textbf{Acknowledgements:} Min Jin Chong's work at UIUC based on work supported in part by NSF Grant 1718221 and in part by ONR MURI Award N00014-16-1-2007.

{\small
\bibliographystyle{ieee_fullname}
\bibliography{egbib}
}

\newpage

\section{Supplementary Material for {\em Retrieve in Style: Unsupervised Facial Feature Transfer and Retrieval}}  %

\maketitle
\thispagestyle{empty}

\section*{Overview}

Even though RIS framework is built upon a pretrained StyleGAN which generates fake images, we focus on applying RIS to real images in the main paper. For completeness, we show RIS on fake images in the supplementary. We further provide more results that could not fit in the main paper due to space constraints. In particular, we offer deeper discussion on these aspects:
\begin{compactenum}
    \item We elaborate the {\bf submembership analysis} on the contribution scores $\mathbf{M}_k$ \cite{collins2020editing} with respect to overlapping channels across different clusters.
    \item We show {\bf latent interpolation} between the source and reference images, verifying the smooth transition for the facial feature transfer.
    \item We enumerate the {\bf attribute classifier accuracy} available in the CelebA attribute dataset and their correspondence to describe facial features, confirming that the accuracy of retrieval performance is meaningful.
\end{compactenum}

\section{Submemberships}

\begin{figure}[thb]
    \centering
    \includegraphics[width=1\linewidth]{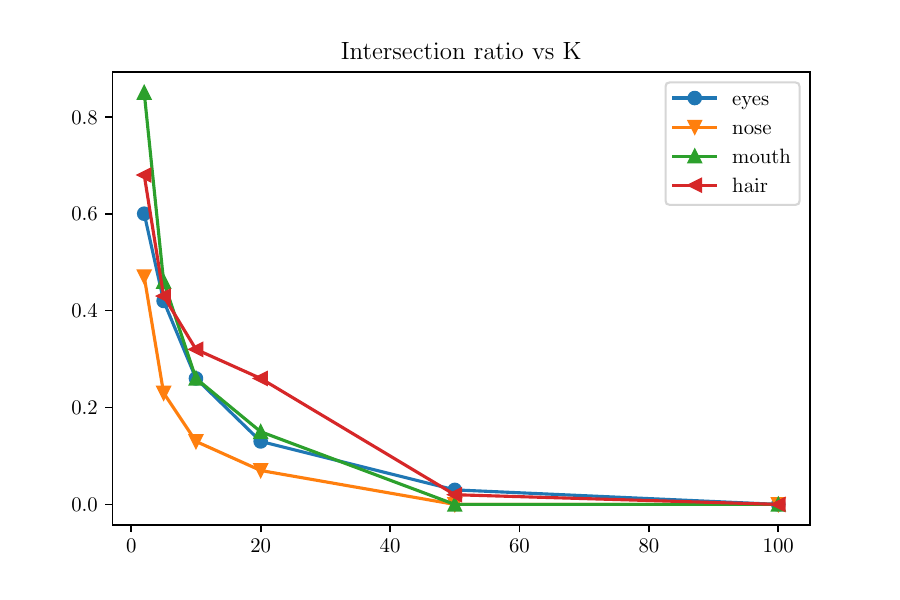}
    \caption{{\bf Intersection Ratio}: 
    This figure shows the intersection ratio ({\bf y-axis}) computed against $K$, the number of clusters ({\bf x-axis}). 
    The common channels shared by all clusters decrease as the number of clusters increase. This means that for the same facial feature, images do not share the same contributing channels, validating the ``submembership'' effect discussed in Sec.~{\color{red}{3.1}} of the main paper.
    }
    \label{fig:intersection}
\end{figure}

A central claim to the proposed method, Retrieve in Style (RIS), is the concept of submemberships, \ie, highly contributing channels that vary from image to image. 
In order to validate the existence of submemberships as discussed in Sec.~{\color{red}{3.1}} of the main paper, we conducted the following experiment. 
We generated $N=5000$ images and computed their $\MM_k$ for a particular feature $k$. 
Then, we performed spherical $K=\{2,5,10,20,50,100\}$-way clustering and averaged each cluster's $\MM_k$. 
Denote $\MM^i_k$ as the average contribution score of feature $k$ for all images belonging to cluster $i$. 
With a slight abuse of notation, we obtain:
\begin{equation}
    \ZZ^i_k = \operatorname{argsort}_n \MM^i_k,
\end{equation}
where $\operatorname{argsort}_n$ is a sorting operator that returns the indices of the top $n$ leading values of $\MM^i_k$ ($n=100$ in our case).
That is, $\ZZ^i_k$ represents the set of top-$n$ most contributing channel for feature $k$ cluster $i$. 
Suppose that there exists a universal $\MM_k$ for all images, $\ZZ^i_k$ should have a high degree of intersection since the important channels for all clusters should be the same. 
We thus define an {\em intersection ratio} as the number of channels common in $\ZZ^i_k$ divided by the $n$. 
From Fig.~\ref{fig:intersection}, the intersection ratio for different features progressively decreases as the number of clusters increases. 
This means that as the clusters get more specific, the number of overlapping channels decreases, validating our hypothesis on submemberships.

\section{Interpolation of Transfers}
\begin{figure*}[thb]
    \centering
    \includegraphics[width=1\linewidth]{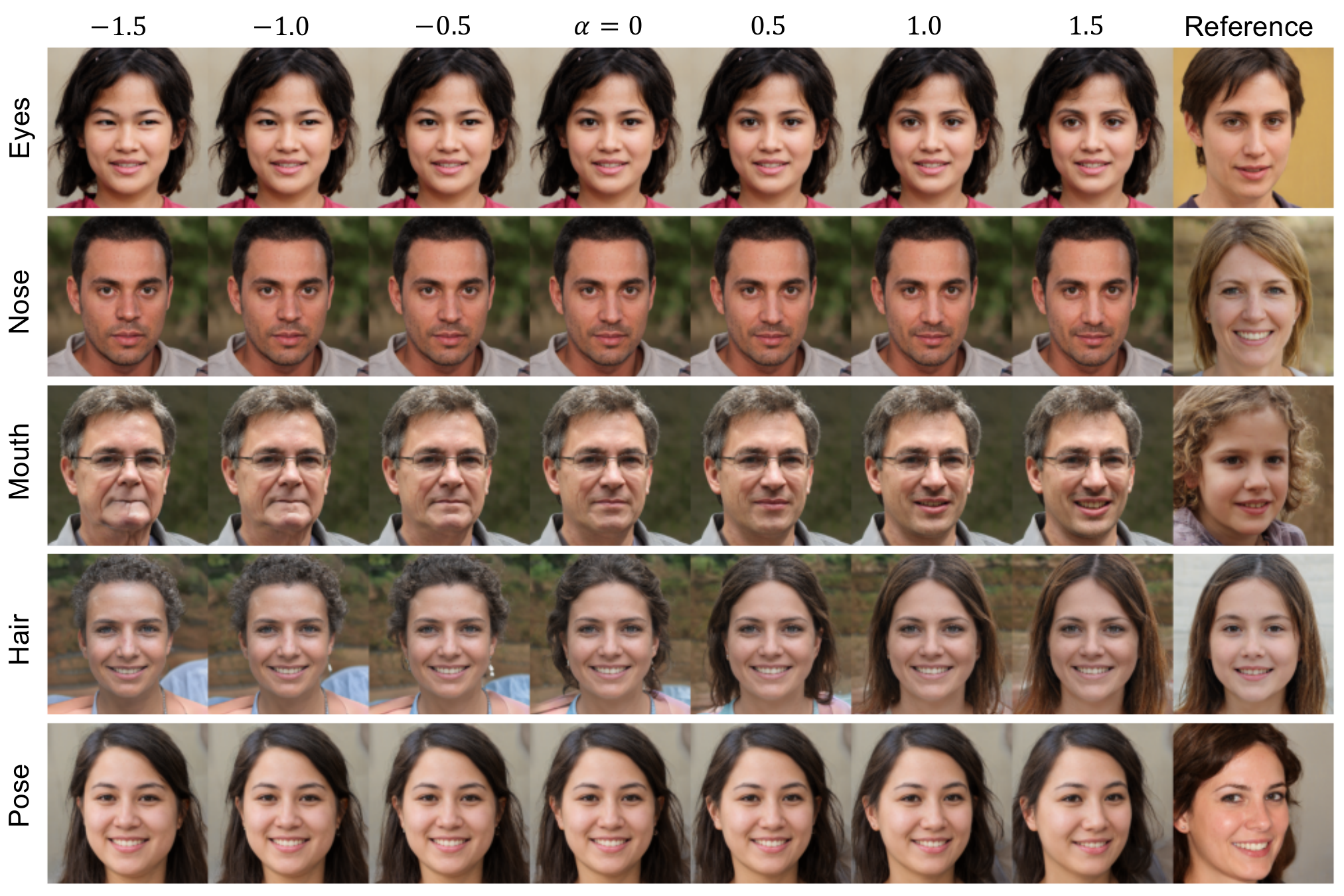}
    \caption{We scale $\qq_k$ according to different $\alpha$ to allow interpolation between the source image (the left most column) and the reference image (the right most column) on a particular facial feature.
    With the side-by-side comparisons with different $\alpha$, we observe that RIS is able to produce smooth and realistic transitions between the transfers.
    The larger value the $\alpha$, the closer the facial features are similar to the reference images.
    Note that hair and pose transfers were not shown possible in the state-of-the-art EIS \cite{collins2020editing}.
    }
    \label{fig:interpolation}
\end{figure*}

In this section, we show that the proposed RIS allows smooth interpolations for facial feature transfers for generated images, in addition to the results shown in Fig. 5 of the original paper. Fig.~\ref{fig:interpolation} shows natural and smooth transition for our interpolation on the target facial features, \ie, eyes, nose, mouth, hair, and pose.
Note that hair and pose transfers were not shown possible in the state-of-the-art EIS approach \cite{collins2020editing}.

\begin{figure*}[thb]
    \centering
    \includegraphics[width=1\linewidth]{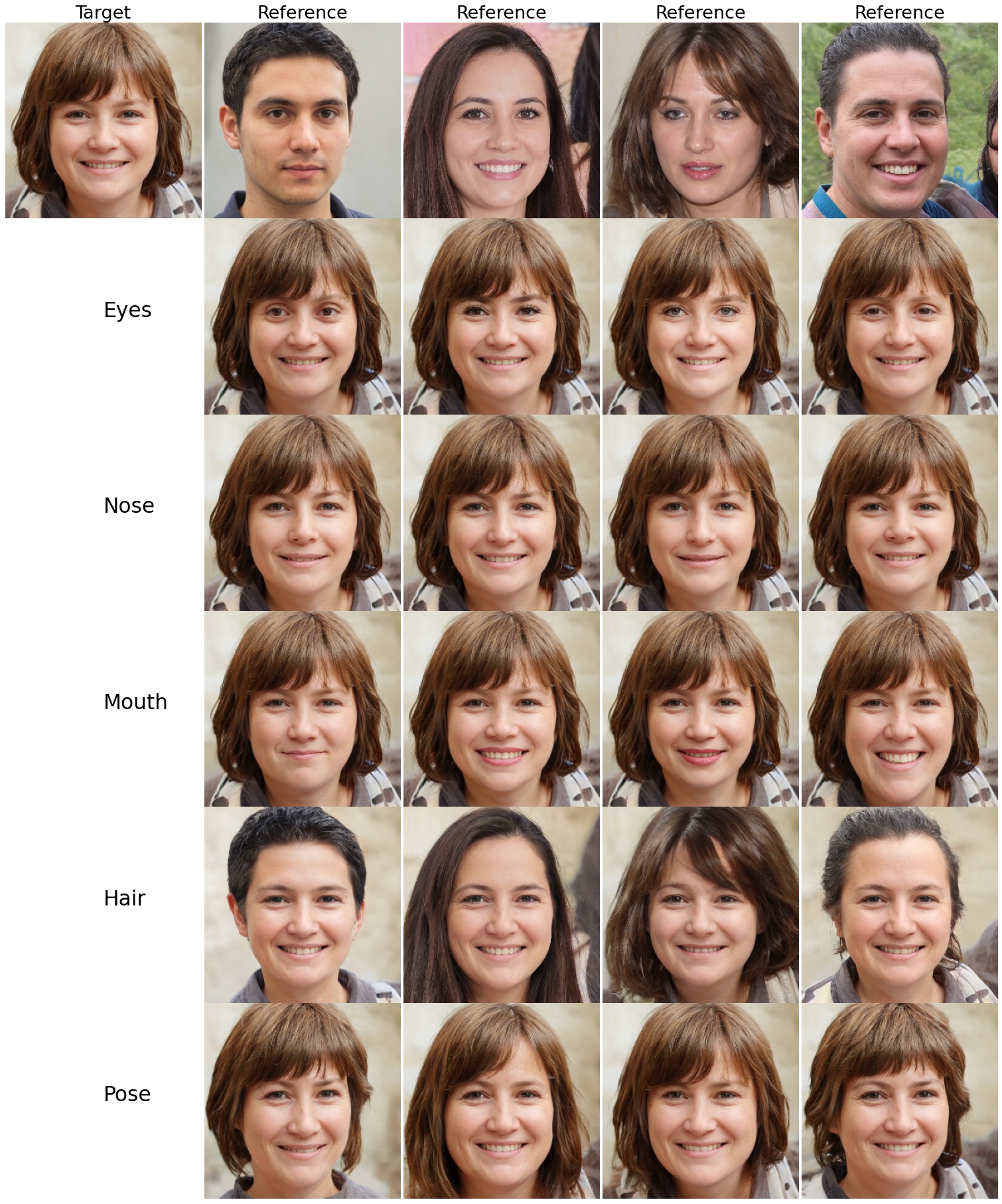}
    \caption{Results of facial feature transfer on generated images.}
    \label{fig:supp1}
\end{figure*}

\begin{figure*}[thb]
    \centering
    \includegraphics[width=1\linewidth]{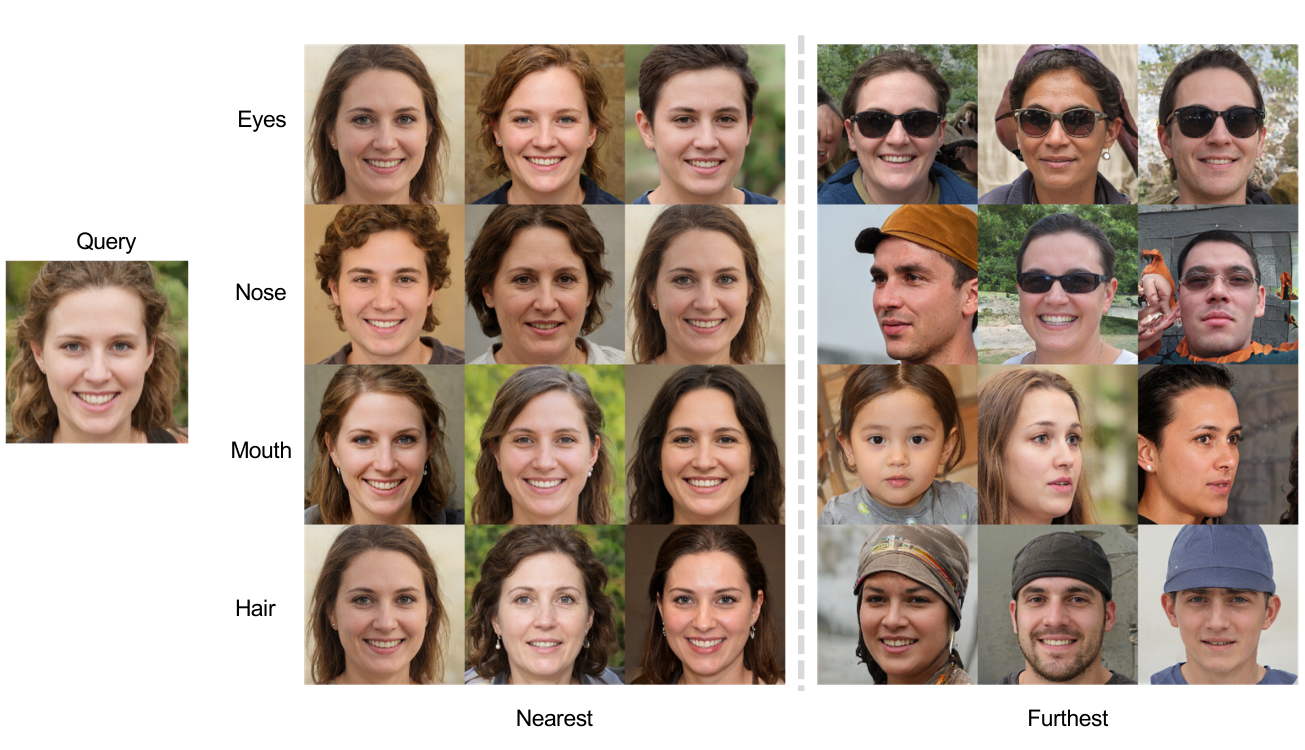}
    \caption{Results of retrieval on generated images.}
    \label{fig:supp2}
\end{figure*}

{\bf More results:}
Similar to the figures shown for facial feature transfer and retrieval as in the main paper, Figs.~\ref{fig:supp1} and \ref{fig:supp2} provide more examples for facial feature transfer retrieval, respectively on generated images.


\section{Attribute Classifier for AMS score}

\begin{figure*}
    \includegraphics[width=\textwidth]{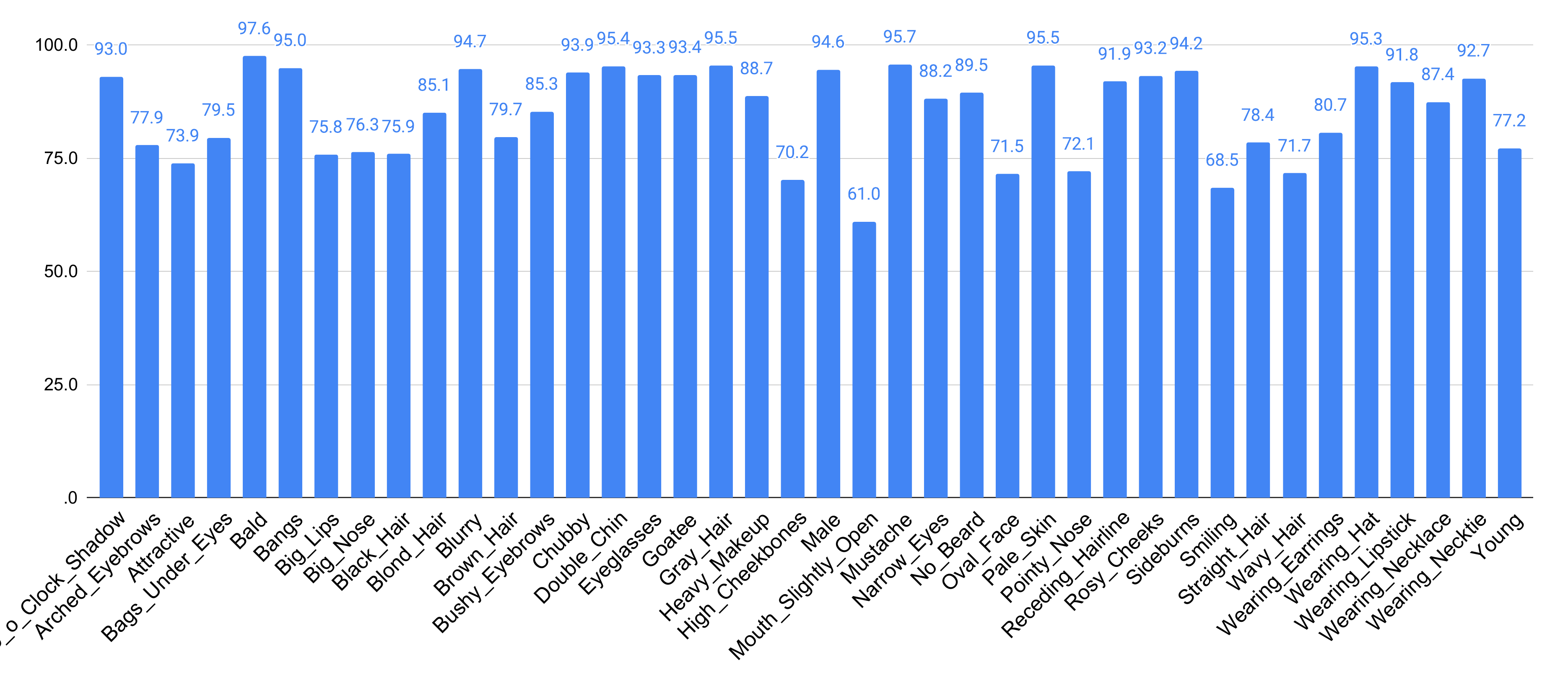}
    \caption{Accuracy on 40 CelebA attributes (in \%).}
    \label{fig:att_acc}
\end{figure*}

In this section, we provide details about attribute classifiers that were used to evaluate our Attribute Matching Score (AMS) in Sec.~{\color{red}{4.2}} of the original paper.
In particular, we pretrained a attribute classifier based on 40 attributes on the CelebA dataset \cite{liu2015faceattributes}. 
Subsets of features were manually selected to associate attributes with the facial features that the proposed method attempts to retrieve.
Table \ref{tab:attributes} shows the full list of binary attributes for each facial feature.
For completeness, Fig.~\ref{fig:att_acc} illustrates the accuracy of each of the 40 attributes of our pretrained model, with an average of 85.27\% overall accuracy.

\begin{table*}[b]
    \centering
    \small
    \begin{tabular}{ll}
        \toprule
        {\bf Facial Feature} & {\bf CelebA Attributes} \\
        \midrule
        Eyes & Arched Eyebrows, Bags Under Eyes, Bushy Eyebrows, Narrow Eyes. \\
        Nose & Big Nose, Pointy Nose. \\
        Mouth & 5 of Clock Shadow, Big Lips, Goatee, Mouth Slightly Open, Mustache, No Beard, Smiling, Wearing Lipstick. \\
        Hair & Bald, Bangs, Black Hair, Blond Hair, Brown Hair, Gray Hair, Receding Hairline, Sideburns, Straight Hair, Wavy Hair. \\
        \bottomrule
    \end{tabular}
    \caption{The relationship between facial features and CelebA attributes that we used to evaluate Attribute Matching Score (AMS) in Sec.~{\color{red}{4.4}} in the main paper. }
    \label{tab:attributes}
\end{table*}

\section{TRSI-IoU metric}
The goal of TRSI-IoU is to measure how disentangled the facial feature representations are, and not the accuracy of retrieval (which is evaluated by Attribute Matching Score). 
For the task of fine-grained feature retrieval, it is pertinent to sufficiently disentangle the feature representations, \ie, the retrieval results of eyes should not predict the retrieval results of nose. 
In an extreme case where features are fully entangled, the identities retrieved across different features become the same. This task is then trivially reduced to the conventional identity retrieval, a simpler and well-researched task compared to our goal of fine-grained feature retrieval.
We observe that EIS retrieves the same images and identities for different features (as shown in Fig.~7(a) and (b) for EIS), which signify {\em significant entanglement} between facial features. 
TRSI-IoU is thus introduced to quantify this entanglement. 
The combination of AMS and TRSI-IoU gives a comprehensive evaluation of both accuracy and entanglement.

\section{Inference speed}
For both EIS and RIS, we perform 100 inference runs (includes both computing $\MM$ and generating the edited image), and compute the mean and standard deviation of the runs on a single Titan Xp GPU. 
Measured in seconds, we observe for EIS: 0.0394$\pm$0.00289, for RIS: 0.234$\pm$0.00633. 
Although computing instance-level $\MM$ adds $\sim$0.2s latency, we believe RIS remains suitable for real world applications. 
Computing $\MM$ for a dataset of 50K images for retrieval takes less than 10 minutes on a single Titan Xp GPU (avg 0.12s per image).

\section{Effects of noise input}
\begin{figure}[ht!]
    \centering
    \includegraphics[width=1\linewidth]{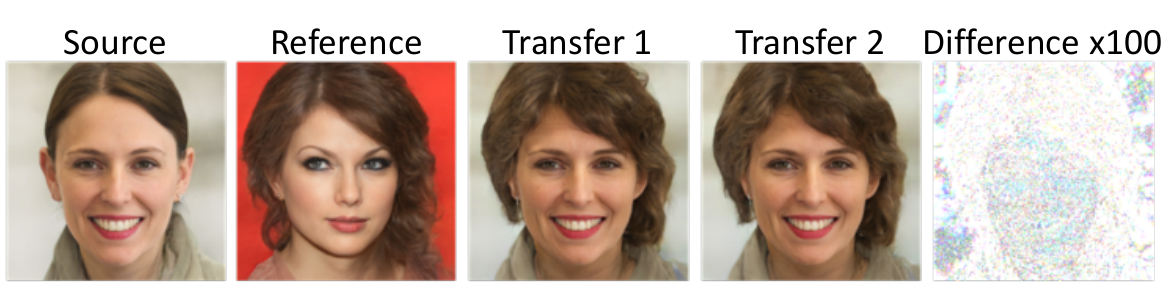}
    \caption{
    {\bf Hair transfer with random noise input}: The effect of noise is negligible to our results even with 100x magnification.
    }
    \label{fig:random_noise}
\end{figure}

In all experiments, we fix the noise input to prevent variations caused by the random noise. 
We perform an experiment showcasing the effect of varied noise input on RIS, as shown in Fig.~\ref{fig:random_noise}. 
From the absolute difference between different random runs, we observe that their delta is negligible. 

\end{document}